\documentclass[journal]{IEEEtran}
\ifCLASSINFOpdf
\else
\fi

\usepackage[utf8]{inputenc} 
\usepackage[T1]{fontenc}    
\usepackage{hyperref}       
\usepackage{url}            
\usepackage{booktabs}       
\usepackage{amsfonts}       
\usepackage{nicefrac}       
\usepackage{microtype}      
\usepackage{lipsum}
\usepackage{graphicx}
\usepackage{xcolor}
\usepackage{soul}
\usepackage[
singlelinecheck=false 
]{caption}

\usepackage{cite}  

\makeatletter
\def\@citex[#1]#2{\leavevmode
\let\@citea\@empty
\@cite{\@for\@citeb:=#2\do
{\@citea\def\@citea{,\penalty\@m\ }%
\edef\@citeb{\expandafter\@firstofone\@citeb\@empty}%
\if@filesw\immediate\write\@auxout{\string\citation{\@citeb}}\fi
\@ifundefined{b@\@citeb}{\hbox{\reset@font\bfseries ?}%
\G@refundefinedtrue
\@latex@warning
{Citation `\@citeb' on page \thepage \space undefined}}%
{\@cite@ofmt{\csname b@\@citeb\endcsname}}}}{#1}}
\makeatother

\begin{document}
%
\title{A COMPUTER VISION APPROACH TO COMBAT LYME DISEASE}
%
%
%

\author{Sina~Akbarian,
        Tania~Cawston,
        Laurent~Moreno,
        Samir~Patel,
        Vanessa~Allen,
        and~Elham~Dolatabadi
\thanks{S. Akbarian is with Public Health Ontario, University of Toronto, and Vector Institute.}
\thanks{T. Cawston is with Public Health Ontario}
\thanks{L. Moreno is with Vector Institute.}
\thanks{S. Patel and V. Allen are with Public Health Ontario and University of Toronto.}
\thanks{E. Dolatabadi is with University of Toronto and Vector Institute. (\emph{e-mail: elham.dolatabadi@mail.utoronto.ca})}}

\maketitle

\begin{abstract}
Lyme disease is an infectious disease transmitted to humans by a bite from an infected Ixodes species (blacklegged ticks). It is one of the fastest growing vector-borne illness in North America and is expanding its geographic footprint. Lyme disease treatment is time-sensitive, and can be cured by administering an antibiotic (prophylaxis) to the patient within 72 hours after a tick bite by the Ixodes species. However, the laboratory-based identification of each tick that might carry the bacteria is time-consuming and labour intensive and cannot meet the maximum turn-around-time of 72 hours for an effective treatment. Early identification of blacklegged ticks using computer vision technologies is a potential solution in promptly identifying a tick and administering prophylaxis within a crucial window period. In this work, we build an automated detection tool that can differentiate blacklegged ticks from other ticks species using advanced deep learning and computer vision approaches. We demonstrate the classification of tick species using Convolution Neural Network (CNN) models, trained end-to-end from tick images directly. Advanced knowledge transfer techniques within teacher-student learning frameworks are adopted to improve the performance of classification of tick species. Our best CNN model achieves 92\% accuracy on test set. The tool can be integrated with the geography of exposure to determine the risk of Lyme disease infection and need for prophylaxis treatment. 

\end{abstract}

\begin{IEEEkeywords}
Lyme disease, Ixodes, Computer Vision, Knowledge Transfer, Convolution Neural Network.
\end{IEEEkeywords}

%
\IEEEpeerreviewmaketitle

\section{Introduction}

Lyme disease is the most common tick-borne disease in North America and is caused by a bacteria called \textit{Borrelia burgdorferi}~\cite{American-tick,smith2011lyme}. The bacteria is primarily transmitted to humans through the bite of an infected tick called \textit{Ixodes scapularis} (or “blacklegged" tick)~\cite{cook2015lyme,parola2001ticks,yssouf2013matrix}. Following a tick bite, an administration of antibiotic prophylaxis is highly effective in curing Lyme disease if given within 72 hours ~\cite{Canada-tick,American-tick}. Rapid diagnostic tests are therefore critical in this public health response.

The current approaches for tick identification are either through using the morphology of taxonomic keys~\cite{parola2001ticks} or molecular methods (e.g. gene sequencing)~\cite{yssouf2013matrix} in a laboratory. However, these methods are time-consuming, expensive to be implemented, and require laboratory facilities and a trained technician. Consequently, laboratory based identification of tick service does not provide fast enough results in deciding whether antibiotic prophylaxis is warranted or not in a patient with a tick bite. 

To address this challenge, an initial digital solution (a mobile application) was developed by the Bishops University and the Public Health Agency of Canada to facilitate the submission of ticks to the lab facilities. Users of the mobile app, called eTick\footnote{\url{https://etick.ca}}, were able to submit the picture of a tick to be reviewed by an entomologist. Although this technology can mobilize the healthcare resources and provides information about the prevalence of the tick species in the region, it still requires manual review by experts, which is subjective and time-consuming.

Early identification of blacklegged ticks using computer vision technologies is a potential solution in mitigating all the risks related to the existing process of tick identification and can significantly prevent development of Lyme disease. Recent advances in computer vision and deep learning have inspired research and development in clinical decision making efforts. These approaches have led to the development of novel and robust diagnostic tools; ie. for medical imaging~\cite{irvin_chexpert:_2019, rajpurkar_chexnet:_2017,rajpurkar_deep_2018,wang_chestx-ray8:_2017, CheXclusion_2020,yao_learning_2017,esteva_dermatologist-level_2017,diabetic2018,Alzheimer2018,IVF2018}, infectious diseases ~\cite{tek2009computer,mohanty2016using}, and sleep apnea monitoring~\cite{Sina}. They also have the potential to revolutionize population health and infectious disease diagnostics.

The combination of the computer vision algorithms and geography location of exposure will help users manage tick bites in real-time. Moreover, rapid identification of the ticks is also improving Lyme disease surveillance as it captures user’s information about potential risk areas. In this work, we deployed advanced computer vision models to build a classifier that can automatically identify blacklegged ticks from other tick species using thousands of tick images. In addition, in order to facilitate adoption and future potential implementation of this technology into a real-life environment, we developed a web application for external validation of the model in the identification of blacklegged ticks. Our proposed solution has the potential to be integrated into a tool positioned to assist clinical decision-making. The tool would enable clinicians to identify the tick species and consider the risk of infection. It does also support patients to seek medical care if they are at risk of developing Lyme disease in real-time. 

This paper is organized as follows: Section II summarizes related works. Section III describes the data sets used in this work, our proposed approach in developing convolutional neural network (CNN) models for automated detection of ticks, and the web application for user interface (UI). Section IV presents our results. Section V discusses the takeaways and concludes the work.

\section{Related Work}
Deep learning algorithms powered by advances in computation and very large datasets have proven to exceed human performance in object detection~\cite{Jang2019LearningWA}. Transfer learning became integral to computer vision tasks as the application of deep learning became ubiquitous for real world problems where there is not a sufficient volume of training data~\cite{Jang2019LearningWA}. In the context of transfer learning, a model which is already pre-trained on a large image dataset (such as ImageNet~\cite{imagenet_cvpr09}) is fine-tuned on a target dataset (e.g. medical images) with minimal modifications where most of the parameters remain frozen during training~\cite{Pan2009}. A pre-trained network trained on large datasets with thousands of classes, various illumination conditions, different backgrounds, and orientation is a powerful tool to extract features from a small amount of training data~\cite{Power-pretrain}. Using transfer learning, the network keeps its ability to extract low-level features acquired from the source domain and discovers how to combine these features to detect complex patterns on the target domain~\cite{knoll2019assessment}. However, Raghu {\em et al.}~\cite{raghu_transfusion:_2019}, indicated that big and small CNN architectures could have similar performances when the training dataset is small. Moreover, Jang {\em et al.}~\cite{Jang2019LearningWA} also showed that transfer learning did not enhance the result especially if the target and source dataset are remarkably distinct, ie. ImageNet and medical imaging. 

In order to address shortcomings with basic transfer learning, several advanced approaches were proposed, including Knowledge Distillation (KD) in Neural Network, which is a knowledge transfer between a teacher and a student network~\cite{bucilu2006model,hinton2015distilling}. Using this approach, a student network could imitate the soft output of a more extensive teacher network or ensemble of networks. Label Smoothing Regularization (LSR)~\cite{LSR} is an extension of KD used as a regularization method. LSR converts one-hot encoded labels (hard labels) to soft labels with a mixture of uniform distribution. Attention transfer (AT) proposed by Zagoruyko {\em et al.}~\cite{attention} is another teacher-student training scheme for knowledge transfer using teacher’s feature maps to guide the learning of the student. Using this approach, given the special attention maps of a teacher network, the student network is trained to learn the exact behavior of the teacher network by trying to replicate it's output at a layer receiving attention from the teacher. The number of AT and the position of the layers depend on whether low-, mid-, and high-level representation information are required. 

In this paper, we adopted AT~\cite{attention,akbarian2020evaluating} and LSR~\cite{LSR} to improve the performance of our deep learning classifier due to the small size of our tick dataset - on the order of thousands of tick images. In addition to performance improvement, both AT and LSR are also utilized for model compression, whereby a small network (student) is taught by a larger trained neural network (teacher)~\cite{hinton2015distilling,wang2020knowledge}. It enables the deployment of CNN models on mobile phones or website applications, which is the long term goal of this work. This paper’s findings are the first step toward a smartphone application for the early diagnosis of Lyme disease.

\section{Methods}
In the following section, we describe the CNN classifiers we built to detect blacklegged ticks versus other tick species. We also present the web application we developed to deploy the CNN models.

\subsection{Dataset Description}
\begin{figure}
  \centering
  \includegraphics[scale=0.32]{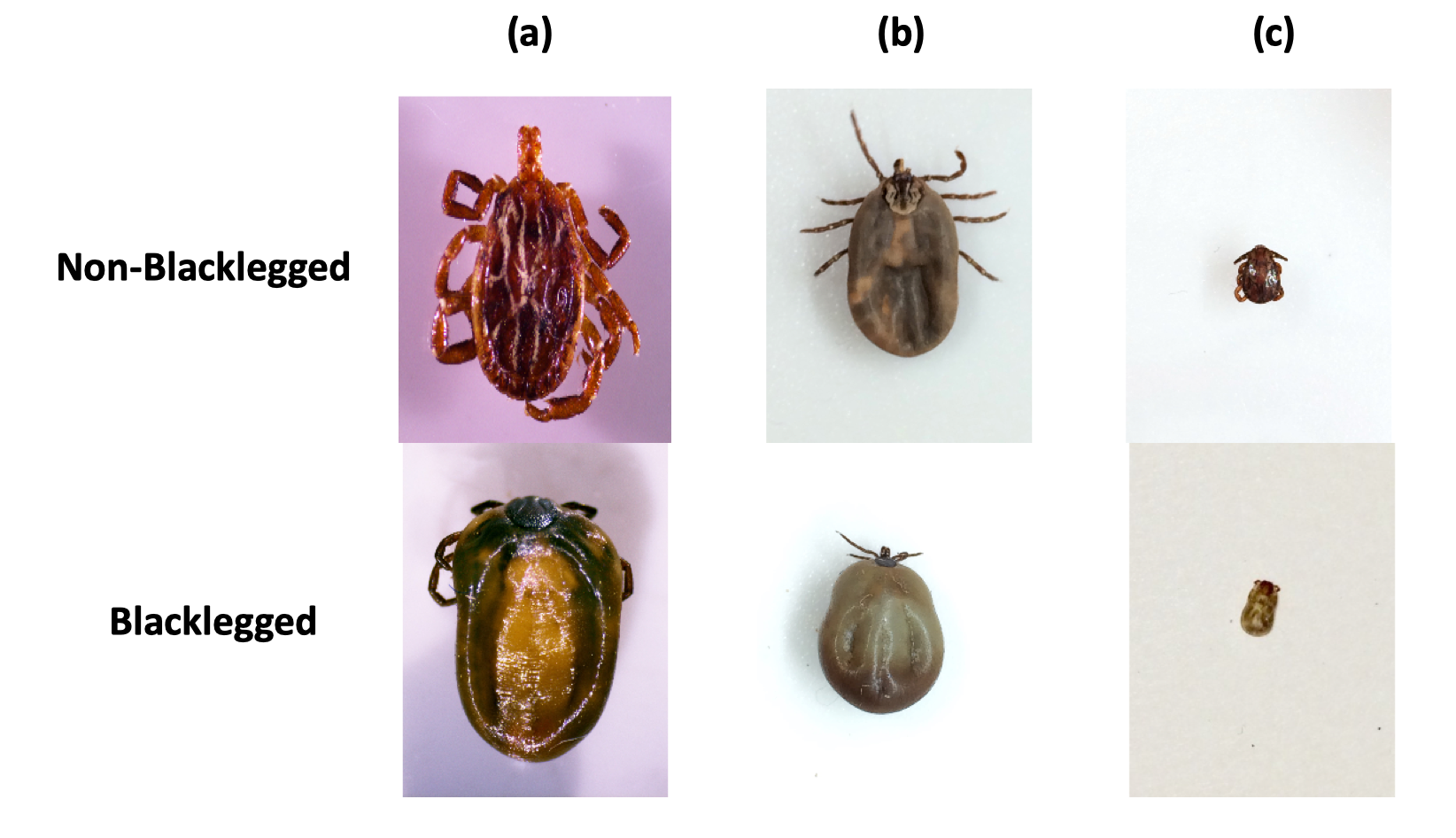}
  \caption{\small (a) High-resolution microscopic images, (b) Mobile phone images of fully engorged ticks, and (c) Mobile phone images of unfed ticks. Fully and slightly engorged ticks can triple in volume when filled with blood.}
\label{tick}
\end{figure}

Our tick dataset was collected by Public Health Ontario, which includes images of blacklegged and other tick species (such as American Dog tick and Lone Star tick) from May to November 2019. All ticks were received by the Public Health Ontario Labs and identified at the Sault Ste Marie location under a stereomicroscope. Given the long term goal of the project which is development of a smartphone application, camera phones (iPhone 5s, 6) were used for image acquisition. The phones were mounted 8 centimeters above the ticks which were placed on a white paper. In total, 12,588 images were captured – 2 per tick, one dorsal and one ventral. Moreover, in order to improve the quality of our dataset, 1000 high-resolution tick images were taken with a camera mounted on the lab stereomicroscope. Our image dataset included 6,294 distinct ticks, of which 41\% were blacklegged, and 59\% were non-blacklegged ticks. A spread of fully engorged, slightly engorged, unfed, and Nymph types were included in our dataset. All tick images were manually annotated by an expert at Public Health Ontario. Fig.~\ref{tick} shows a sample of tick images in the dataset.

\subsection{CNN classifiers}
\label{model}
In this paper, we conducted different frameworks to train our CNN classifiers:
\begin{enumerate}
    \item training the CNN models from scratch with random initialization (all layers were unfrozen during training),
    \item  knowledge transfer from CNN networks pre-trained on ImageNet.
\end{enumerate}
For the knowledge transfer, we focused on three training strategies: (i) transfer learning from ImageNet, (ii) AT, and (iii) AT combined with LSR (AT + LSR). For AT setting, the teacher networks were the Inception-Resnet~\cite{Inception-resnet} already pre-trained on ImageNet, and student networks were lighter CNN model.\\

\textbf{Lighter CNN model}: Lighter CNN model in this work comprised 7 convolution layers followed by a dropout or batch normalization. In addition, average pooling layers were used to reduce the number of parameters. In total, the network had 13 layers with 5,350,633 trainable parameters out of 5,352,041 parameters (more details of the network is shown in Appendix A).\\

\begin{figure*}[h!]
  \centering
  \includegraphics[scale=0.6]{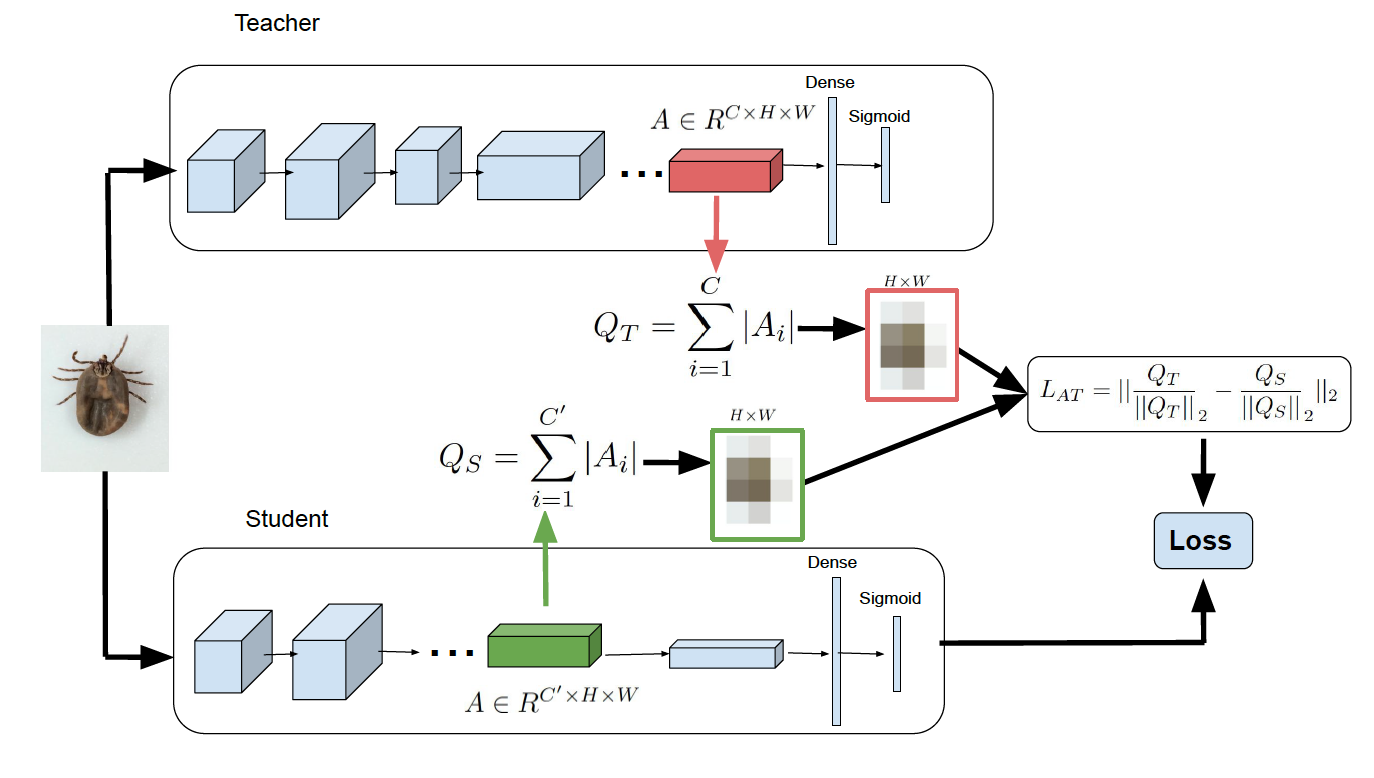}
  \caption{\small An overview of the Attention Transfer (AT) loss in a teacher-student learning setting. The spatial attention map is constructed by taking the sum of absolute values of a layer’s 3D tensor, A, across the channel dimension. In this setting, knowledge is transferred from the last layer of the teacher network to the one before the last layer of the student network. In the shown example, the spatial attention map $Q\in R^{H\times W}$ is 8$\times$8 and teacher's (C) and student's (C') channel dimensions are 1536 and 32, respectively.}
\label{scheme}
\end{figure*}

\textbf{Attention Transfer (AT)}: Following the work of Zagoruyko {\em et al.}~\cite{attention}, we built an activation based AT to transfer knowledge from the last layer of the teacher network (Inception-Resnet) to the one before the last layer of the student network (lighter CNN) as shown in Fig.~\ref{scheme}. The knowledge to be transferred in our setting is a spatial attention map, constructed by taking the sum of absolute values of a layer's 3D tensor $A\in R^{C\times H\times W}$ across the channel dimension:

\begin{equation}
Q=\sum\limits_{i=1}^{C}{|A_i|},
\end{equation}

Where $C$, $H$, and $W$ are channel dimension, height, and width of a CNN layer's tensor A, respectively. The spatial attention map, $Q$, is therefore a 2D tensor $Q\in R^{H\times W}$. Using $l_2$ normalization, we calculated AT loss between the teacher's and student's spatial attention map of the same resolution (same $H$ and $W$) as follows:

\begin{equation}
L_{AT}=||{\frac{Q_T}{||Q_T||}_2 -\frac{Q_S}{||Q_S||}_2}||_2,
\end{equation}

Where $Q_S$ and $Q_T$ are the vectorized form of student's and teacher's spatial attention maps. The overall approach is shown in Fig.~\ref{scheme}.\\

\textbf{Label Smoothing Regularization (LSR)}: In this work, we made use of LSR as a regularization technique to smooth the loss function. For this approach, we trained two student networks where one of the students, student$_1$, was trained on a subset of training data using AT loss. After student$_1$ was trained, it was used to generate soft labels for the entire training data as follows: 

\begin{enumerate}
    \item For correctly classified images, the network produced class probabilities by converting the logits, $\theta_{i}, i\in\{0,1\}$, computed for each class, into a probability $p_{i} = \frac{1}{1+exp^{-\theta_i/T}}$, as suggested in~\cite{KD-hint}. $T$ is a temperature where a higher value for $T$ produces a softer probability distribution over classes.
    \item For incorrectly classified images, the network replaced class probabilities, $p_{i}$, with a constant probability sampled from a uniform distribution. In this work, we chose to replace the predicted probabilities for true classes with 0.6.
\end{enumerate}
The second student network, student$_2$, is therefore trained with the following loss function, which is a weighted combination of AT and LSR:

\begin{equation}
L_{tot}= -\frac{1}{\beta}_{1} \sum\limits_{i=0}^{1}{(p_i\log{q_i})}+ \frac{1}{\beta}_{2}L_{AT}
\end{equation}
\label{eq:loss}
where $p_i$  is the soft label produced by student$_1$, $q_i$ is the output probability predicted by student$_2$, and $\beta_{1}$ and $\beta_{2}$ are the weights balancing attention loss and cross-entropy loss. 

\subsection{Web application development}
\begin{figure}[h!]
  \centering
  \includegraphics[scale=0.25]{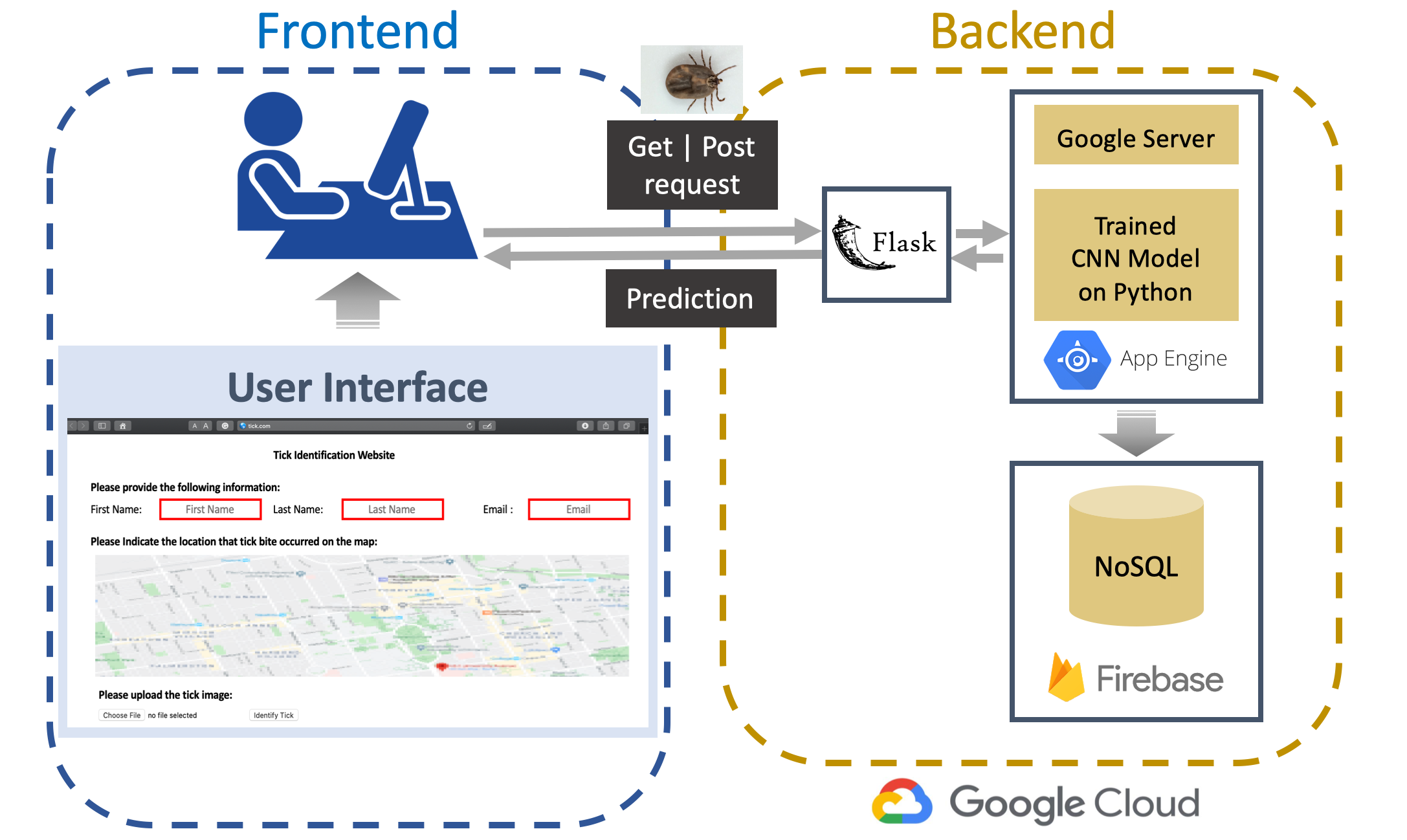}
  \caption{\small The system architecture of deploying our CNN model on the web application for early identification of blacklegged ticks. For the frontend, HTML (HyperText Markup Language) and CSS (Cascading Style Sheets) were used to create the user interface (UI). On the backend, Python Flask application was developed to handle the get and post requests between UI and compute engine. Our trained CNN model was deployed on the app engine of the google cloud platform. The users' data were stored in the Firebase Realtime Database (NoSQL) as JSON and synchronized in real-time to every connected user.}
\label{UI}
\end{figure}

As a second step toward our main objective, we created a web application that was shared internally with Public Health Ontario lab technicians for external validation of the model in the identification of blacklegged ticks. Using the web application, the lab technician can upload the image of a tick taken by a cell phone and receive feedback from the platform in less than a minute. It also captures the geolocation of the exposure and pairs it with public health data, enabling the assessment of the risk of Lyme disease infection and the need for prophylaxis treatment. Fig.~\ref{UI} shows the end-to-end web application deployment of the CNN model. The uploaded data is processed in the backend on the compute engine of the google cloud and results will be provided to users in less than a minute.

\section{Result}
In this section, the classification results obtained by applying different CNN models on the tick dataset are presented. 

For model development and evaluation, our dataset was divided into a train/test split with a ratio of 11/1 without any overlap. Therefore, 12,554 images (41\% blacklegged) were used for the training set and 1034 (41\% blacklegged) were used for the test set. The training data was augmented with random rotation of $0^{\circ}$-$360^{\circ}$, horizontal flip, vertical flip, and zoom range of 0.5-2x. Adam was used to optimize the loss function in all of the experiments. K-fold (k=3) cross validation was used for model evaluation and hyper-parameter tuning on the validation set.
The input image sizes for the lighter CNN model and Inception-Resnet network were $300 \times 300$ and  $299 \times 299$, respectively. The lighter CNN model was trained for maximum 256 epochs with an initial learning rate of $1e^{-3}$ and a batch size of 64. For the AT approach, the classification loss was the combination of $L_{AT}$ and binary cross entropy loss. For the AT + LSR approach, the loss parameters (eq.3), including $\beta_{1}$, $\beta_{2}$, and T, were set to be 1, 2, and 5, respectively.

Table~\ref{tbl:exp1} reports the results of our first experiment, where the performance of training the lighter CNN and standard Inception-Resnet~\cite{Inception-resnet} models are compared. The lighter CNN was trained from scratch with random initialization while the standard Inception-Resnet was trained through transfer learning using ImageNet weights in addition to the random initialization. For the transfer learning, we conducted two tests where in one setting all layers were unfrozen to be trained translating to 53 \emph{m} trainable parameters, and in the other setting, we froze all the network layers except the last five (5) layers translating to only 4.5 \emph{m} trainable parameters. As the results of our first experiment (Table 1) indicate, training the Inception-Resnet model either from scratch with random initialization or from ImageNet pre-trained weights without any frozen layers have the highest performances on accuracy, area under the ROC curve, and area under the precision-recall curve. We can also observe from the results that the lighter CNN obtained comparable results to both of Inception-Resnet CNN models. So, the initial layers of the network should be included and unfrozen during training the model as fine-tuning just the last layers of the CNN network on tick images perform very poorly. 

\begin{table*}[ht]

\caption{\small The performance of using different strategies including the network size and initialization for training CNN classifiers to differentiate between the two common tick species; blacklegged vs dog ticks. The best performances per each column are in bold and the second best scores are underlined. ROC-AUC is the area under the ROC curve and PR-AUC is the area under the precision recall curve. Regardless of initialization, CNN models with larger number of trainable parameters perform better on tick dataset. The CNN classifier performs very poorly if the initial layers are fixed during training. \textbf{*} Only the last 5 layers of the Inception-Resnet were unfrozen for retraining while the rest of the CNN in the Table were trained from scratch without any frozen layers. }
\centering
\begin{center}

\begin{tabular}{lccccc}
\toprule
Model & Initialization & \# Trainable Parameters & Accuracy & ROC-AUC & PR-AUC \\
\hline
Lighter CNN\ & Random & 5.3 $m$&91.68 $\pm$ 0.25 & 97.55 $\pm$ 0.34& 95.43 $\pm$ 0.46 \\
Inception-Resnet & Random & 53 $m$ & \textbf{92.04} $\pm$ 0.48& \textbf{98.52} $\pm$ 0.28 & \textbf{96.80} $\pm$ 0.99 \\
Inception-Resnet$^*$\ & ImageNet & 4.5 $m$& 42.10 $\pm$ 0.37 & 57.85 $\pm$ 0.08 & 47.96 $\pm$ 0.20\\
Inception-Resnet & ImageNet & 53 $m$& \underline{91.75} $\pm$ 0.06& \underline{98.51} $\pm$ 0.38 & \underline{96.77} $\pm$ 0.89\\
\bottomrule
\end{tabular}
\end{center}

\label{tbl:exp1}
\vspace{-10pt}
\end{table*}

In our second experiment, we examined AT and AT + LSR techniques from a teacher network to a student network as shown in Table~\ref{tbl:exp2}. As explained in section~\ref{model}, two student networks were trained for AT + LSR where one student network generates soft labels. As the results indicate both AT and AT + LSR models performed the same across all measures. Comparing all CNN models from Table~\ref{tbl:exp1} and Table~\ref{tbl:exp2} together, we can observe that knowledge transfer mechanism (Table~\ref{tbl:exp2}) outperforms training CNN from scratch with random initialization (Table~\ref{tbl:exp1}) based on test accuracy measures. However, all models achieve comparable performance on the area under the ROC curve and the area under the precision recall curve. 

\begin{table}[t]
\centering
\caption{\small The performance of using Attention Transfer (AT) and Attention Transfer with Label Smoothing Regularizer (AT + LSR) for classification of blacklegged ticks versus other tick specious. Teachers are Inception-Resnet pre-trained on ImageNet, and students are lighter CNN model with 5.3 $m$ trainable parameters. The best performances per each column are in bold. Smoothing the loss function through LSR approach makes the CNN model perform slightly better on accuracy measure.}
\begin{center}

\begin{tabular}{lcccc}
\toprule
Model & Accuracy & ROC-AUC & PR-AUC \\
\hline
AT  & 91.20 $\pm$ 0.33 & \textbf{97.70} $\pm$ 0.29 & \textbf{96.69} $\pm$ 0.08\\
AT + LSR & \textbf{92.55} $\pm$ 0.39 & 97.32 $\pm$ 0.32 & 96.17 $\pm$ 0.05\\
\bottomrule
\end{tabular}
\end{center}

\label{tbl:exp2}
\vspace{-10pt}
\end{table}

\begin{figure}[h!]
  \centering
  \includegraphics[scale=0.35]{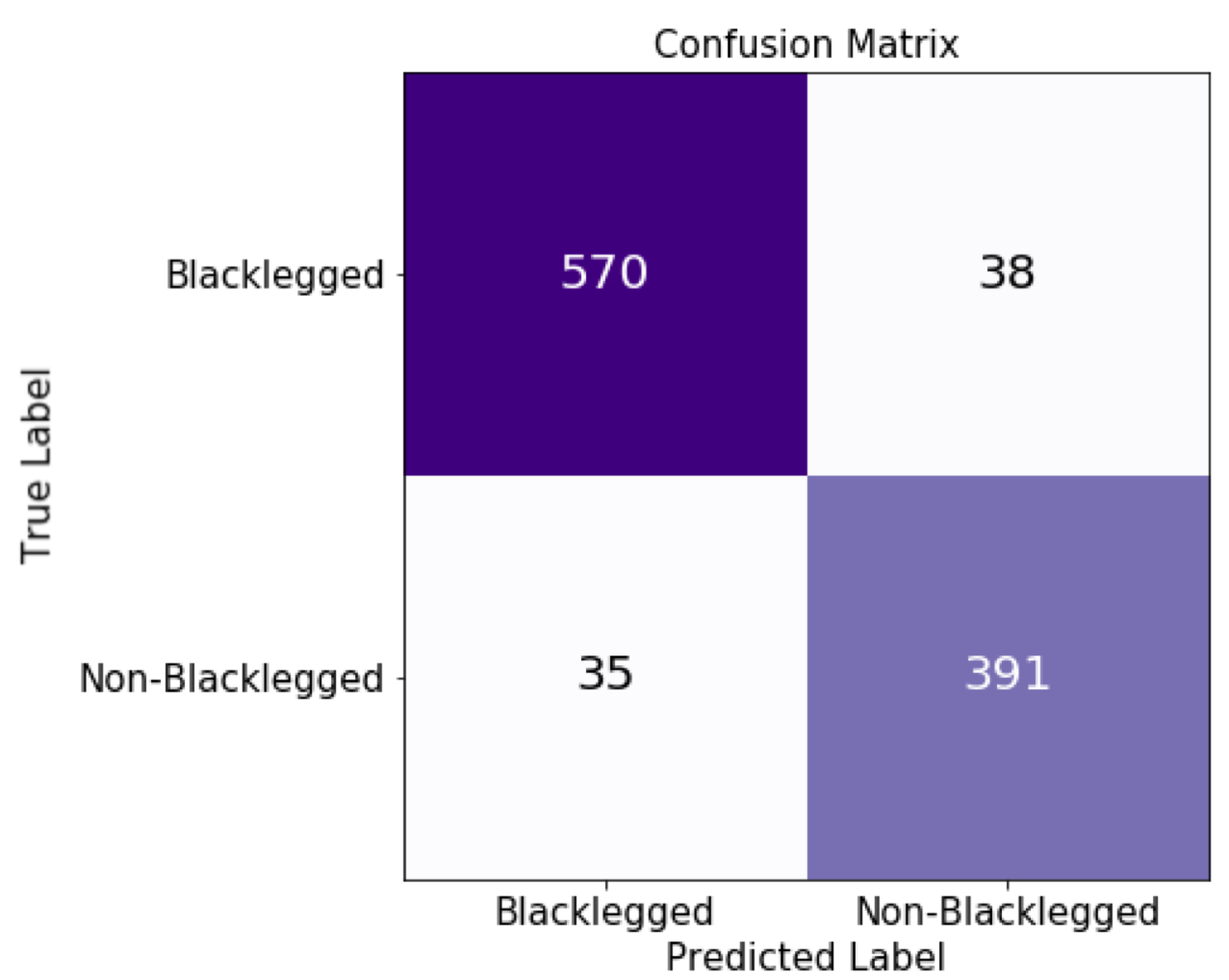}
  \caption{\small Confusion matrix for the best performing model which is the combination of attention transfer and label smoothing regularization on the test set.}
\label{conf}
\end{figure}

We selected AT + LSR as the best performing model to be deployed on the web application given the accuracy measures shown in Table~\ref{tbl:exp1} and~\ref{tbl:exp2}, and the confusion matrix of the best model (AT + LSR) is shown in Fig.~\ref{conf}.

\section{Discussion}
Towards simplifying and automating diagnosis of Lyme disease, we proposed and presented an alternative tool to that of existing laboratory approaches for tick identification. This work has been created based on clinicians’ priorities and following extensive consultations with microbiologists and infectious disease specialists at Public Health Ontario. Our partners were therefore involved in all stages of development and validation of our proposed tool.
In this preliminary effort, we demonstrated the potential usefulness of advanced CNN models for classification of blacklegged ticks versus other tick species. Our tick dataset included several noisy blurry images due to the presence of unfed and nymph ticks. The white background of images could also bias the CNN models and affect their generalization performance. We, therefore, adopted advanced transfer learning and knowledge transfer approaches in order to minimize the effect of our dataset issues on the accuracy and generalization of our CNN model. Our best classification model was able to identify blacklegged ticks with 92\% accuracy using AT and LSR techniques. In this setting, a small CNN model receives knowledge from a large CNN model and learns to behave like a large network during classification. As shown in our sets of experiments, not only does this model outperforms other models in terms of accuracy, but also it has the potential to be deployed on small devices such as mobile phones due to the small size of the network (5.3 trainable parameters).

Building the deep learning computer vision models was our first step toward a smartphone application for early diagnosis of Lyme disease. As a second step toward our primary goal, we created a web application that was shared internally with Public Health  Ontario lab technicians for external validation of the model in the identification of blacklegged ticks. We are still in the process of evaluating the web application in close partnership with public health experts and after completion of the validation trial, our future work includes public release of the web application. We also plan to develop a mobile app in similar settings where users can upload the tick picture and the geographical location of the tick bite and receive a response in real-time. Users' responses would include not only the type of tick but also the likelihood of the tick carrying the bacteria that causes Lyme disease using the area of distribution of the ticks causing the infection provided by Public  Health  Ontario. As more users utilize the app, more data will be collected which will improve the accuracy of the classifier. In addition, the surveillance of tick species distribution will also be improved.  

This work provides some evidence that advanced deep learning technologies hold great promise for early identification of Lyme disease. However, how these technologies will eventually be adapted and incorporated into an affordable, sensitive, specific, and user-friendly tool for end-users require to be explored. We hope this work will be helpful to those interested in advancing and adopting deep learning models in the field of infectious disease diagnostics.

\section*{Acknowledgment}
We would like to acknowledge Vector Institute and also its high performance computing platforms made available for conducting this work. We would also like to acknowledge Public Health Ontario lab at the Sault Ste Marie location and Hiba Hussain for assistance with data acquisition and coordination. This work has been funded by Vector Institute and Public Health Ontario through Pathfinder projects.

\ifCLASSOPTIONcaptionsoff
  \newpage
\fi



%
\bibliographystyle{IEEEtran}  
\bibliography{references}

\onecolumn
\newpage
\begin{appendices}
\appendices
\section{}
In the appendix, we present supplementary details about lighter CNN model as listed in Table~\ref{tappendix}. The network has 13 layers with 5,350,633 trainable parameters composed of convolution layers and average pooling layers. 


\begin{table*}[h]
\centering
\caption{The architecture of lighter CNN.}
\begin{tabular}{|c|c|c|c|c|}
\hline
			
Layer & Number of filters, n & Size/stride & Activation function & Output size\\
\hline\hline
Input & N/A &  N/A & N/A & 3 x 300 x 300\\
\hline
Convolutional & 64 & 8/2 & N/A & 64 x 147 x 147\\
\hline
Batch normalization & N/A & N/A & Leaky Relu & 64 x 147 x 147\\
\hline
Convolutional & 128 & 8/1 & N/A & 128 x 140 x 140\\
\hline
Batch normalization & N/A & N/A & Relu & 128 x 140 x 140\\
\hline
Average pool & N/A & 4/2 & N/A & 128 x 69 x 69\\
\hline
Dropout & N/A & N/A & N/A & 128 x 69 x 69\\
\hline
Convolutional & 256 & 8/1 & N/A & 256 x 62 x 62\\
\hline
Batch normalization & N/A & N/A & Relu &256 x 62 x 62\\
\hline
Convolutional & 128 & 8/1 & N/A & 128 x 55 x 55\\
\hline
Batch normalization & N/A & N/A & Relu & 128 x 55 x 55\\
\hline
Average pool & N/A & 4/2 & N/A & 128 x 26 x 26\\
\hline
Dropout & N/A & N/A & N/A & 128 x 26 x 26\\
\hline
Convolutional & 64 & 8/1 & N/A & 64 x 19 x 19\\
\hline
Batch normalization & N/A & N/A & Relu &64 x 19 x 19\\
\hline
Convolutional & 32 & 5/2 & N/A & 32 x 8 x 8\\
\hline
Batch normalization & N/A & N/A & Relu & 32 x 8 x 8\\
\hline
Convolutional & 32 & 5/1 & N/A & 32 x 4 x 4\\
\hline
Batch normalization & N/A & N/A & Relu & 32 x 4 x 4\\
\hline\hline
Flatten & N/A & N/A & N/A & 512\\
\hline\hline
Fully connected & 32 & 512 x 32 & Relu & 32\\
\hline
Fully connected & 4 & 32 x 4 & Relu & 4\\
\hline
Output layer&N/A & 4 x 1 & Sigmoid & 1\\
\hline
\end{tabular}
\label{tappendix}
\end{table*}
\end{appendices}




%






\end{document}